# Interval Valued Trapezoidal Neutrosophic Set for Prioritization of Non-functional Requirements

Kiran Khatter, *Department of Computer Science, BML Munjal University*

*Abstract*—This paper discusses the trapezoidal fuzzy number(TrFN); Interval-valued intuitionistic fuzzy number(IVIFN); neutrosophic set and its operational laws; and, trapezoidal neutrosophic set(TrNS) and its operational laws. Based on the combination of IVIFN and TrNS, an Interval Valued Trapezoidal Neutrosophic Set (IVTrNS) is proposed followed by its operational laws. The paper also presents the score and accuracy functions for the proposed Interval Valued Trapezoidal Neutrosophic Number (IVTrNN). Then, an interval valued trapezoidal neutrosophic weighted arithmetic averaging (IVTrNWAA) operator is introduced to combine the trapezoidal information which is neutrosophic and in the unit interval of real numbers. Finally, a method is developed to handle the problems in the multi attribute decision making(MADM) environment using IVTrNWAA operator followed by a numerical example of NFRs prioritization to illustrate the relevance of the developed method.

## 1. INTRODUCTION

Zadeh developed the fuzzy set theory [1] to deal the impreciseness, incompleteness and uncertainty in the information. Later, Zadeh [2] in 1975 proposed the interval valued fuzzy sets(IVFS) if grade of membership is uncertain and cannot be expressed in terms of a crisp value.
 Atanassov [3] extended the fuzzy set theory and developed an intuitionistic fuzzy set(IFS) [3][4][5]. Various researchers have explored the use of IFSs in MCDM situations[6][7][8], stock market prediction [9] and medical diagnosis[10].
Liu and Yuan [11] combined the concept of IFS and triangular fuzzy numbers (TFN), and introduced the triangular intuitionistic fuzzy sets (TIFS). Further, Atanassov and Gargov [12] combined the IFS and IVFS, and introduced the interval valued intuitionistic fuzzy set (IVIFS). Further, the use of IVIFS was demonstrated in MADM [13] and multi attribute group decision making(MAGDM) [14] situations. Wang [15] proposed the weighted geometric and hybrid geometric operators using triangular intuitionistic fuzzy sets. Further, he applied both the operators to handle MAGDM problems. Wei et al. [16] proposed an induced ordered weighted geometric operator on the basis of Fuzzy number intuitionistic fuzzy numbers and introduced an approach based on the proposed operator to solve group decision making problems. Ye [17] extended the TIFS and proposed the trapezoidal intuitionistic fuzzy set (TrIFS) for representing the membership and non-membership values in the form of a trapezoid. Smarandache [18] extended the concept of classic, fuzzy and IFS, and proposed the neutrosophic set(NS) to deal imprecise, incomplete and uncertain information. Later, A variation of a NS i.e. single-valued neutrosophic set(SVNS) is proposed which can be applied in real world scenarios [19]. Jun Ye [20] introduced the TrNS as an extension of trapezoidal fuzzy numbers (TrFN) and SVNS. He also introduced weighted arithmetic and geometric averaging operator based on the trapezoidal neutrosophic number. Further, using these operators, he introduced a method to handle MADM problems.  As discussed, various methods have been proposed by the researchers based on IVIFS, TrIFS, and TrNS set to handle inconsistency, impreciseness, uncertainty, incompleteness and indeterminacy in the information where information is either (1) neutrosophic and can be represented in the form of a trapezoid (2) or the information is intuitionistic fuzzy and in the unit interval of real numbers and can be represented in the form of a triangle/trapezoid. But the proposed methodology handles the information which is neutrosophic in nature and in the unit interval of real numbers and can be represented in the form of a trapezoid or a triangle.

Thus the paper proposes an interval valued trapezoidal neutrosophic set (IVTrNS) based on the combination of IVIFN and TrNS. The paper also introduces the operational laws for IVTrNN. Further an interval valued trapezoidal neutrosophic weighted arithmetic averaging (IVTrNWAA) operator is introduced to combine the trapezoidal information which is neutrosophic and in the unit interval of real numbers. Finally, a method is developed to handle the problems in the MADM environment using IVTrNWAA operator followed by a numerical example of NFRs prioritization to illustrate the relevance of the developed method. Remaining sections of the paper are organized as follows: Section 2 presents the concept of non-functional requirements (NFRs) and discusses the interdependencies among NFRs. Section 3 introduces the preliminaries related to IVIFN and TrNS. Section 4 proposes an IVTrNS as a generalization of TrNS and IVIFN and introduces some operational laws of IVTrNS. In section 5, the score and accuracy functions of the proposed IVTrNN are proposed. In Section 6, the IVTrNWAA operator is proposed to aggregate the interval valued trapezoidal neutrosophic information. Section 7 develops a MADM method using the proposed IVTrNWAA operator, score and accuracy functions. In Section 8, NFRs prioritization is performed using interval valued trapezoidal neutrosophic information to illustrate the relevance of the developed method. Section 9 discusses the conclusion remarks.

## 2. NON-FUNCTIONAL REQUIREMENTS

Requirements Analysis is a most important aspect in developing the quality software because "if requirements are not correct, errors caused by insufficient requirement analysis affect the design and implementation phases of software development life cycle". These errors account for a large number of unprofitable software products because repairing these errors is highly expensive and time consuming process. Thus, the phase which decides the successful completion of a project is Requirement Analysis [21]. Poor analysis of requirements affects the budget of development and it gets increased by 70-85% due to revisiting all the phases of software development in order to accommodate the revised requirements [22]. Requirements are classified as Functional and Non-functional requirements: Functional Requirement (FR) allows the user to operate the software for the desired function. It defines what software is supposed to do. Non-functional Requirement (NFR) is a restriction on the requirement which must be accommodated during the design phase of software. It specifies the criteria to judge the functionality of a system whereas functional requirement concerns the specific functionality of a system. For example, "The system shall check authenticity of the users before allowing access to the data", is a functional requirement but under what constraints this requirement is going to be satisfied will be known as NFR. For example, "The system shall perform authentication within 5 seconds" is a NFR. The NFRs concept has been widely investigated by various software researchers. One of the key challenges in handling non-functional requirements is that there is no proper definition of NFRs. There are different interpretations on non-functional requirements as reported by various researchers in their work. Few researchers treated NFRs as the restrictions on software development processes while others considered non-functional requirements as quality attributes that stakeholders be concerned about. There is difference in the concepts considered in the definitions of NFRs [23] [24]. FR refers the behavioural aspects of a system [25] whereas NFR refers the non-behavioural traits of a system [26]. FR means "what" the system must function whereas NFR means "how" the system must perform [27].

The IEEE states the Non-functional Requirements as: "Non-functional Requirement in software system engineering is a software requirement that describes not what the software will do, but how the software will do it. Non-functional requirements are difficult to test; therefore, they are usually evaluated subjectively."

The most important aspect for extracting and analyzing FR and NFR is the interactions with various stakeholders such as end users, customers, developers, analyst, designers etc. The aim of Requirements Engineering is to ensure that system must meet all stakeholders' needs and expectations by properly understanding, extracting, specifying and validating the FRs and NFRs. Since most of the software systems have long lifecycles and long development cycles, it is evident that business needs and requirements will change with the time and as a result, system has to be evolved with the changes. This requires the synchronization among all stakeholders' needs and requirements to keep the development activities consistent. Since all stakeholders come from different background and have different expectations from the software, during the requirements elicitation process, stakeholders disagree over the interpretation of software need and its intended use. This leads to conflict among requirements and prioritization is needed to resolve that conflict. There are many different interpretations of stakeholders' perspectives and their interrelationships. Users are concerned about the functionality and usability of the software; customers are concerned about the desired quality of the software and cost of developing the software whereas developers focus on risk management and maintainability of the software. These different perspectives generally overlap each other and give rise to conflicts, in such cases, which perspective is to be considered at first becomes a question. Some requirements may not be compatible with other requirements and cannot be achieved together giving rise to mutually exclusive conflicts. If requirements are not fully consistent with each other, it will lead to partially interfering conflicts. In fact, requirements are never satisfied in isolation and normally satisfaction of one requirement may affect sometimes negatively to the satisfaction of another. Satisfaction of Security requirement might vote for the usage of biometric or two factor authentication, but two factor authentication might affect the level of Usability requirement. Therefore, a methodology is needed to maintain multiple prospective and knowledge about their inter-relationships and conflicts simultaneously. These different perspectives on the system not only need to be combined at the starting phase of development, in fact, this harmonization is a continuous activity during the whole life-cycle. This can be possible only by prioritization of the requirements to enable the selection of the optimal requirement.

## 3. PRELIMINARIES

### 3.1 Fuzzy Set

There are many programming languages (C, Java, COBOL etc.) which are appropriate for representing the mathematical models or logical reasoning in software systems, but mathematical models lack in incorporating or considering the uncertainty, human thinking and ability to take a decision. Software based systems use Boolean logic for decision making whereas human beings use imprecise and indefinite expression such as excellent, very high, high, good or very poor to make a decision. In this context, Zadeh[1] proposed the framework of Fuzzy Set mathematically to work in uncertain and ambiguous situations to solve the problem with incomplete information and poorly defined concepts such as low reliability, good performance, high maintenance etc. [28]

*Definition:* Let $X$ be universe of discourse. A fuzzy set $A$ in $X$ is a set of ordered pairs $A = \{\langle x, \mu_A(x)\rangle : x \in X\}$ where each element of $X$ is mapped to $[0,1]$ by $\mu_A: X \to [0,1]$. The fuzzy set considers single real value $\mu_A(x) \in [0,1]$ for representing the grade of membership of $A$ on $X$. Thus grade of non-membership of $x$ into $A$ is $1 - \mu_A(x)$. The membership function helps to represent the fuzzy set graphically. The $x$ and $y$ axis refers to the universe of discourse and the degree of membership respectively in the $[0,1]$. Zadeh[1] has defined various membership functions such as Triangular, Singleton, $L$ Function etc. Depending upon type of membership function, different fuzzy sets can be obtained. The membership of triangular function [Figure 1] defined by a lower and upper limit ($a$ and $c$ respectively), and a value $m$ where $a < m < c$ is as follows:

$$\mu_A(x) = \begin{cases} \dfrac{x-a}{m-a} & a < x < m \\ 1 & x = m \\ \dfrac{c-x}{c-m} & m < x < c \\ 0 & x \geq c \end{cases} \quad (1)$$

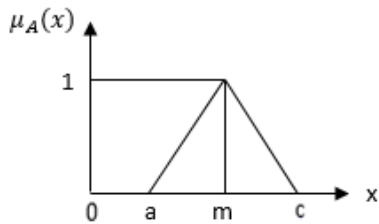

Figure 1: Membership of a Triangular function

The Fuzzy set has become a dominant area of research and a powerful tool for the evaluation of real world scenarios. Soon after the fuzzy set introduced by Zadeh, various extensions were proposed by researchers and these extensions were L-fuzzy sets [29], IVFSs [2][30], rough sets [31] and IFSs [2].

### *3.2 Interval Valued Fuzzy Set(IVFS)*

Since Fuzzy Sets consider single value for representing the grade of membership and sometimes grade of membership is uncertain and it is hard to specify by a crisp value. For example, it is difficult for a software developer to exactly quantify his opinion about the reliability of software, therefore it is appropriate to represent the degree of certainty by an interval. In order to consider the uncertainty of grade of membership, Zadeh [2] introduced the concept of IVFS which uses an interval value to represent the grade of membership of fuzzy set $A$. Let $R = [a^-, a^+]$, $a^-, a^+ \in [0,1]$, $a^- \leq a^+$ then a mapping $A: X \to [0,1]$ is known as an Interval Valued Fuzzy Sets [32].

*Definition*[33][34]**:** An interval $A(x) = [A^L(x), A^U(x)]$ represents the IVFS $A$ defined on universe $X$ where $A^L$ is lower fuzzy set ($A^L: X \to [0,1]$) and $A^U$ is upper fuzzy set ($A^U: X \to [0,1]$):

$$A = \{(x, [A^L(x), A^U(x)]) : x \in X\}, \quad 0 \leq A^L(x) \leq A^U(x) \leq 1$$

### *3.3 Intuitionistic Fuzzy Set(IFS) and Interval Valued Intuitionistic Fuzzy Set(IVIFS)*

In some cases where human judgement is uncertain and it needs to be incorporated in the solution of a problem, then we should consider both the memberships: Truth and Falsity which was not taken into account in the fuzzy sets and IVFS. Atanassov [2] developed the IFS which considers both the truth-membership and falsity membership. Let $X$ be a universe of discourse. Then an IFS $A$ in $X$ is defined as $A = \{\langle x, \mu_A(x), v_A(x)\rangle : x \in X\}$ where $0 \leq \mu_A(x) + v_A(x) \leq 1$. The function $\mu_A: X \to [0,1]$ represents the degree of membership function of element $x$ whereas $v_A: X \to [0,1]$ represents the degree of non-membership of $x$. In IFS, indeterminacy is $1 - \mu_A(x) - v_A(x)$ by default and it is not quantified explicitly, however we can define a function $\pi_A: X \to [0,1]$ by $\pi_A(x): 1 - \mu_A(x) - v_A(x)$ to represent the degree of indeterminacy [35]. For example, when we ask the software developer about the reliability of software, he may claim that the possibility that system is reliable is 0.4 and the system is not reliable is 0.5 and the degree that he is uncertain about the reliability of the system is 0.1 [19].

In case of an IVIFS, $\mu_A(x)$ and $v_A(x)$ will hold the values in the interval such that $\mu_A(x) = [\mu_A^-(x), \mu_A^+(x)]$ and $v_A(x) = [v_A^-(x), v_A^+(x)]$ with the condition $0 \leq \mu_A^+(x) + v_A^+(x) \leq 1$.

### *3.4 Neutrosophic Set(NS)*

In neutrosophic set(NS), indeterminacy is computed separately and degree of indeterminacy was introduced as an independent component by F. Smarandache [18]. He defined neutrosophy as "a branch of philosophy which studies the origin, nature and scope of neutralities, as well as their interactions with different ideational spectra". Neutrosophic set is based on the concept of classic set, fuzzy set, IVFS, IFS etc. Let $X$ be a universe of discourse. A neutrosophic set $A$ in $X$ is defined as $A = \{\langle x, T_A(x), I_A(x), F_A(x)\rangle : x \in X\}$ where $T_A(x) + I_A(x) + F_A(x) \in\ ]^-0, 1^+[$. $T_A$, $I_A$ and $F_A$ represents the degree of truth, indeterminacy and falsity membership of element $x$ in the set $A$ respectively. $T_A(x), I_A(x)$ and $F_A(x)$ considers the value from subintervals in the real standard/non-standard $]^-0, 1^+[$. It means $T_A: X \to ]^-0, 1^+[$, $I_A: X \to ]^-0, 1^+[$ and $F_A: X \to ]^-0, 1^+[$. There is no restriction on the sum of $T_A(x), I_A(x)$ and $F_A(x)$, thus $^-0 \leq \sup T_A(x) + \sup I_A(x) + \sup F_A(x) \leq 3^+$.

### *3.5 Single Valued Neutrosophic Set(SVNS)*

Since neutrosophic set considers the value from subintervals in the real standard/non-standard $]^-0, 1^+[$ which will be difficult to apply in scientific/engineering applications. Therefore Wang et al. [19] defined SVNS in which truth, indeterminacy and falsity membership functions take the value from real standard $[0,1]$. Let $X$ be a universe of discourse. A SVNS $A$ in $X$ is defined as $A = \{\langle x, T_A(x), I_A(x), F_A(x)\rangle : x \in X\}$ where $T_A(x), I_A(x)$ and $F_A(x) \in [0,1]$. It means $T_A: X \to [0,1]$, $I_A: X \to [0,1]$ and $F_A: X \to [0,1]$. There is no restriction on the sum of $T_A(x), I_A(x)$ and $F_A(x)$, thus $0 \leq T_A(x) + I_A(x) + F_A(x) \leq 3$.

When $X$ is continuous, a SVNS would be[19]:

$$A = \int_X \langle x, T(x), I(x), F(x)\rangle / x : x \in X \qquad (2)$$

In case of X as discrete, a SVNS would be [19]:

$$A = \sum_{i=1}^{n} \langle x, T(x_i), I(x_i), F(x_i)\rangle / x : x \in X \qquad (3)$$

The Internet technology has revolutionized the approach of doing business and targeting the market. Due to growth of bandwidth, cloud services and computer usage, every business has come up with web-interface not only to provide the global reach of their product and services but also to reduce the operational costs and deliver better services to their customers. Due to such reasons, banks offer e-services for customized financial services to fulfil customer needs, and meet the customer preferences and quality expectations. Though Internet technology is helping banks to expand the market across boundary, to offer customized financial products and services and to reduce operational costs, but reliability of transaction processing, usability, performance and transaction security are the critical factors for success. Therefore e-service portal of the bank must address these issues at the early stages of development. All stakeholders (System analyst, Software developer etc.) of the software system are asked to give their opinion on *Reliability* of a transaction processing ($REL$) and transaction *Security* ($SECU$) and it could be the degree of high, degree of indeterminacy (uncertainty) and degree of low. The *reliability* refers to the capability of a software system to maintain its consistent performance for a specified time period under the specified environment and to ensure the reliability, software must be robust, available and recoverable under adversity [36]. Thus *Reliability*, which is a key NFR, of a system further depends upon the availability ($AVAIL$), maintainability ($MAIN$) and recoverability ($RECOV$) NFRs of the system.

Every software system must be available to render service whenever it is needed by the users. Thus software *Availability* indicates the software reliability during operational hours.

*Maintainability* is a group of planned activities that work together in order to prevent the loss of functionality contributing to the reliability of a software system.

*Recoverability* is another key non-functional requirement to evaluate the reliability of a software. In order to measure the downtime that a business can sustain, it is essential to study how often software fails to deliver expected output. Good software architecture specifies the recoverability in time required for maintenance in a service-level agreement. In order to achieve the Recoverability software requirement, regular backup ($BKUP$) and data mirroring ($MIRR$) must be properly implemented so that system is recoverable in case of any disaster or failure [28].

According to Glinz [37], for some requirements, the satisfaction level is either completely satisfied or not satisfied (*discrete requirement*) and for some others, satisfaction level depends upon a range of acceptable behavior (*continuous requirement*). Consider three non-functional requirements Security($SECU$), Recoverability($RECOV$) and *Reliability* ($REL$) which are continuous in nature. Reliability ($REL$) non-functional requirement is further achieved by two NFRs: Availability ($AVAIL$) and *Maintainability* ($MAIN$) which are continuous in nature. Assume that $X = [SECU, MAIN, AVAIL]$ and values of $SECU, MAIN$ and $AVAIL$ are in [0,1]. A SVNS $A$ in $X$ for continuous non-functional requirement (using (2)) is defined as

$$A = \left\{ \int_{SECU} \langle 0.2, 0.3, 0.4 \rangle / SECU, \int_{MAIN} \langle 0.3, 0.5, 0.6 \rangle / MAIN, \int_{AVAIL} \langle 0.5, 0.2, 0.3 \rangle / AVAIL \right\} \quad (4)$$

In order to achieve the Recoverability($RECOV$) software requirement, backup at regular intervals ($BKUP$) and data mirroring ($MIRR$) must be properly implemented so that system is recoverable in case of any disaster or failure. Thus Recoverability non-functional requirement is achieved by two discrete requirements: $BKUP$ and $MIRR$. Assume that $U = [BKUP, MIRR]$ and values of $BKUP$ and $MIRR$ are in [0,1]. A SVNS $B$ in $U$ for discrete requirements (using (3)) is defined as

$$B = \{\langle 0.7, 0.2, 0.2 \rangle / BKUP + \langle 0.4, 0.2, 0.4 \rangle / MIRR\} \quad (5)$$

### 3.6 Interval Valued Neutrosophic Set(IVNS)

Wang et al. [38] defined IVNS which can also be applied in scientific/engineering applications. Let $X$ be a universe of discourse. An IVNS $A$ in $X$ is defined as $A = \{\langle x, T_A(x), I_A(x), F_A(x) \rangle : x \in X\}$ where $T_A(x), I_A(x)$ and $F_A(x)$ are interval truth member function, interval indeterminacy member function and interval falsity membership function respectively.

If $X$ is continuous in nature, then IVNS can be written as [38]:

$$A = \int_x \langle x, T_A(x), I_A(x), F_A(x) \rangle / x : x \in X \quad (6)$$

In case, X is discrete, IVNS would be [38]:

$$A = \sum_{i=1}^{n} \langle x, T_A(x), I_A(x), F_A(x) \rangle / x : x \in X \quad (7)$$

Assume that $X = [SECU, AVAIL]$ and values of $SECU$ and $AVAIL$ are in $[0,1]$. An IVNS $A$ in $X$ for continuous non-functional requirement is defined as

$$\left\{ \begin{array}{l} \int_{SECU} \langle[0.1,0.3],[0.3,0.5],[0.5,0.8]\rangle : SECU, \\ \int_{AVAIL} \langle[0.1,0.4],[0,0.2],[0.5,0.8]\rangle : AVAIL, \end{array} \right\} \quad (8)$$

Assume that $U = [BKUP, MIRR]$ and values of $BKUP$ and $MIRR$ are in $[0,1]$. An IVNS $B$ in $U$ for discrete requirements is defined as

$$B = \{\langle[0.1,0.3],[0,0.2],[0.5,0.7]\rangle : BKUP + \langle[0.2,0.4],[0,0.1],[0.4,0.8]\rangle : MIRR \} \quad (9)$$

### 3.7 Trapezoidal Neutrosophic Set(TrNS)

TrNS extends the concept of TrFN and SVNS [20].

#### 3.7.1 Trapezoidal Fuzzy Number(TrFN)

A TrFN $\tilde{A} = (a, b, c, d)$ is a fuzzy set on $X$ with the membership function $\mu_A(x)$ as follows:

$$\mu_{\tilde{A}}(x) = \begin{cases} 0 & x < a \\ \dfrac{x-a}{b-a} & a \leq x \leq b \\ 1 & b \leq x \leq c \\ \dfrac{d-x}{d-c} & c \leq x \leq d \\ 0 & x > d \end{cases}$$

where $a < b < c < d$.

In a trapezoidal fuzzy number, trapezoid is divided into three parts where first part is a triangle, second part is a rectangle and third part is a triangle respectively [Figure 2]. If $b = c$ then it will be converted to a triangle fuzzy number.

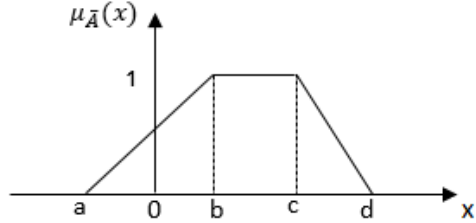

Figure 2: Membership function of Trapezoidal Fuzzy Number

#### 3.7.2 Neutrosophic Number

Let $\tilde{n}$ be a neutrosophic number then $T_{\tilde{n}}(x), I_{\tilde{n}}(x), F_{\tilde{n}}(x)$, representing the truth, indeterminacy and Falsity membership functions respectively, can be represented as follows[39]:

$$T_{\tilde{n}}(x) = \begin{cases} q_{\tilde{n}}(x) & a_1 \leq x < a_2 \\ T_{\tilde{n}} & a_2 \leq x \leq a_3 \\ r_{\tilde{n}}(x) & a_3 < x \leq a_4 \\ 0 & \text{otherwise} \end{cases}$$

$$I_{\tilde{n}}(x) = \begin{cases} k_{\tilde{n}}(x) & b_1 \leq x < b_2 \\ I_{\tilde{n}} & b_2 \leq x \leq b_3 \\ u_{\tilde{n}}(x) & b_3 < x \leq b_4 \\ 1 & \text{otherwise} \end{cases}$$

$$F_{\tilde{n}}(x) = \begin{cases} o_{\tilde{n}}(x) & c_1 \leq x < c_2 \\ I_{\tilde{n}} & c_2 \leq x \leq c_3 \\ z_{\tilde{n}}(x) & c_3 < x \leq c_4 \\ 1 & \text{otherwise} \end{cases}$$

where

$T_{\tilde{n}}, I_{\tilde{n}}$ and $I_{\tilde{n}} \in [0,1]$,

$0 \leq T_{\tilde{n}} + I_{\tilde{n}} + F_{\tilde{n}} \leq 3$

and $a_1, a_2, a_3, a_4, b_1, b_2, b_3, b_4, c_1, c_2, c_3, c_4 \in X$.

The functions $q_{\tilde{n}}, r_{\tilde{n}}, k_{\tilde{n}}, u_{\tilde{n}}, o_{\tilde{n}}, z_{\tilde{n}} : X \to [0,1]$ are called the side of a fuzzy number.

Let
$$A_1 = \{\langle x, T_{A_1}(x), I_{A_1}(x;), F_{A_1}(x)\rangle : x \in X\}$$
and
$$A_2 = \{\langle x, T_{A_2}(x), I_{A_2}(x), F_{A_2}(x)\rangle : x \in X\}$$

be neutrosophic sets, the operational laws are defined as follows [19][39]:

- Addition:
$A_1 \oplus A_2 = \{\langle x, T_{A_1}(x) + T_{A_2}(x) - T_{A_1}(x)T_{A_2}(x), I_{A_1}(x)I_{A_2}(x), F_{A_1}(x)F_{A_2}(x)\rangle : x \in X\}$

- Multiplication:
$A_1 \otimes A_2 = \{\langle x, T_{A_1}(x)T_{A_2}(x), I_{A_1}(x) + I_{A_2}(x) - I_{A_1}(x)I_{A_2}(x), F_{A_1}(x) + F_{A_2}(x) - F_{A_1}(x)F_{A_2}(x)\rangle : x \in X\}$

- Equality:
$A_1 = A_2$ if and only if $A_1 \subseteq A_2$ and $A_2 \subseteq A_1$

- Intersection:
$A_1 \cap A_2 = \{\langle x, T_{A_1}(x) \wedge T_{A_2}(x), I_{A_1}(x) \vee I_{A_2}(x), F_{A_1}(x) \vee F_{A_2}(x)\rangle : x \in X\}$

- Union:
$A_1 \cup A_2 = \{\langle x, T_{A_1}(x) \vee T_{A_2}(x), I_{A_1}(x) \wedge I_{A_2}(x), F_{A_1}(x) \wedge F_{A_2}(x)\rangle : x \in X\}$

- Inclusion:
$A_1 \subseteq A_2$ if and only if $T_{A_1}(x) \leq T_{A_2}(x), I_{A_1}(x) \geq I_{A_2}(x), F_{A_1}(x) \geq F_{A_2}(x)_1$ for any $x$ in $X$

- Complement:
$A_1^C = \{\langle x, F_{A_1}(x), 1 - I_{A_1}(x), T_{A_1}(x)\rangle : x \in X\}$ for any $x$ in $X$

Let $X$ be a universe of discourse. A TrNS $\widetilde{N}$ in $X$ is defined as $\widetilde{N} = \{\langle x, T_{\widetilde{N}}(x), I_{\widetilde{N}}(x), F_{\widetilde{N}}(x)\rangle : x \in X\}$ where $T_{\widetilde{N}}(x) \subset [0,1], I_{\widetilde{N}}(x) \subset [0,1]$ and $F_{\widetilde{N}}(x) \subset [0,1]$ are trapezoidal numbers. It means:
$T_{\widetilde{N}}(x) = \left(t_{\widetilde{N}}^1(x), t_{\widetilde{N}}^2(x), t_{\widetilde{N}}^3(x), t_{\widetilde{N}}^4(x)\right) : X \to [0,1]$, $I_{\widetilde{N}}(x) = \left(i_{\widetilde{N}}^1(x), i_{\widetilde{N}}^2(x), i_{\widetilde{N}}^3(x), i_{\widetilde{N}}^4(x)\right) : X \to [0,1]$ and $F_{\widetilde{N}}(x) = (f_{\widetilde{N}}^1(x), f_{\widetilde{N}}^2(x), f_{\widetilde{N}}^3(x), f_{\widetilde{N}}^4(x)) : X \to [0,1]$.

There is no restriction on the sum of $T_A(x), I_A(x)$ and $F_A(x)$, thus $0 \leq t_{\widetilde{N}}^4(x) + i_{\widetilde{N}}^4(x) + f_{\widetilde{N}}^4(x) \leq 3$.

*Definition:* Let $\tilde{n}$ be a trapezoidal neutrosophic fuzzy number then $T_{\tilde{n}}(x), I_{\tilde{n}}(x)$ and $F_{\tilde{n}}(x)$ can be defined as follows[40]:

$$T_{\tilde{n}}(x) = \begin{cases} \dfrac{x - t_{\tilde{n}}^1}{t_{\tilde{n}}^2 - t_{\tilde{n}}^1} T_{\tilde{n}} & t_{\tilde{n}}^1 \leq x < t_{\tilde{n}}^2 \\ T_{\tilde{n}} & t_{\tilde{n}}^2 \leq x \leq t_{\tilde{n}}^3 \\ \dfrac{t_{\tilde{n}}^4 - x}{t_{\tilde{n}}^4 - t_{\tilde{n}}^3} & t_{\tilde{n}}^3 < x \leq t_{\tilde{n}}^4 \\ 0 & \text{otherwise} \end{cases}$$

$$I_{\tilde{n}}(x) = \begin{cases} \dfrac{i_{\tilde{n}}^2 - x + I_{\tilde{n}}(x - i_{\tilde{n}}^1)}{i_{\tilde{n}}^2 - i_{\tilde{n}}^1} & i_{\tilde{n}}^1 \leq x < i_{\tilde{n}}^2 \\ I_{\tilde{n}} & i_{\tilde{n}}^2 \leq x \leq i_{\tilde{n}}^3 \\ \dfrac{x - i_{\tilde{n}}^3 + I_{\tilde{n}}(i_{\tilde{n}}^4 - x)}{i_{\tilde{n}}^4 - i_{\tilde{n}}^3} & i_{\tilde{n}}^3 < x \leq i_{\tilde{n}}^4 \\ 1 & \text{otherwise} \end{cases}$$

$$F_{\tilde{n}}(x) = \begin{cases} \dfrac{f_{\tilde{n}}^2 - x + F_{\tilde{n}}(x - f_{\tilde{n}}^1)}{f_{\tilde{n}}^2 - f_{\tilde{n}}^1} & f_{\tilde{n}}^1 \leq x < f_{\tilde{n}}^2 \\ F_{\tilde{n}} & f_{\tilde{n}}^2 \leq x \leq f_{\tilde{n}}^3 \\ \dfrac{x - f_{\tilde{n}}^3 + I_{\tilde{n}}(f_{\tilde{n}}^4 - x)}{f_{\tilde{n}}^4 - f_{\tilde{n}}^3} & f_{\tilde{n}}^3 < x \leq f_{\tilde{n}}^4 \\ 1 & \text{otherwise} \end{cases}$$

where $T_{\tilde{n}}, I_{\tilde{n}}, F_{\tilde{n}} \in [0,1]$, $0 \leq T_{\tilde{n}} + I_{\tilde{n}} + F_{\tilde{n}} \leq 3$. It means $t_{\tilde{n}}^1, t_{\tilde{n}}^2, t_{\tilde{n}}^3, t_{\tilde{n}}^4, i_{\tilde{n}}^1, i_{\tilde{n}}^2, i_{\tilde{n}}^3, i_{\tilde{n}}^4, f_{\tilde{n}}^1, f_{\tilde{n}}^2, f_{\tilde{n}}^3, f_{\tilde{n}}^4: X \to [0,1]$, then $\tilde{n} = \langle\, ([t_{\tilde{n}}^1, t_{\tilde{n}}^2, t_{\tilde{n}}^3, t_{\tilde{n}}^4]: T_{\tilde{n}}), ([i_{\tilde{n}}^1, i_{\tilde{n}}^2, i_{\tilde{n}}^3, i_{\tilde{n}}^4]: I_{\tilde{n}}), ([f_{\tilde{n}}^1, f_{\tilde{n}}^2, f_{\tilde{n}}^3, f_{\tilde{n}}^4]: F_{\tilde{n}})\,\rangle$ is a Trapezoidal Neutrosophic number. Let $\tilde{n}_1$ and $\tilde{n}_2$ are two trapezoidal neutrosophic numbers, $\tilde{n}_1 = \langle (a_1, b_1, c_1, d_1), (e_1, f_1, g_1, h_1), (l_1, m_1, n_1, p_1)\rangle$, $\tilde{n}_2 = \langle (a_2, b_2, c_2, d_2), (e_2, f_2, g_2, h_2), (l_2, m_2, n_2, p_2)\rangle$, following operations are defined [20][40]:

- $\tilde{n}_1 \oplus \tilde{n}_2 = \langle (a_1 + a_2 - a_1 a_2, b_1 + b_2 - b_1 b_2, c_1 + c_2 - c_1 c_2, d_1 + d_2 - d_1 d_2), (e_1 e_2, f_1 f_2, g_1 g_2, h_1 h_2), (l_1 l_2, m_1 m_2, n_1 n_2, p_1 p_2)\rangle$

- $\tilde{n}_1 \otimes \tilde{n}_2 = \langle (a_1 a_2, b_1 b_2, c_1 c_2, d_1 d_2), (e_1 + e_2 - e_1 e_2, f_1 + f_2 - f_1 f_2, g_1 + g_2 - g_1 g_2, h_1 + h_2 - h_1 h_2), (l_1 + l_2 - l_1 l_2, m_1 + m_2 - m_1 m_2, n_1 + n_2 - n_1 n_2, p_1 + p_2 - p_1 p_2)\rangle$

- $\lambda \tilde{n}_1 = \langle (1 - (1-a_1)^\lambda, 1 - (1-b_1)^\lambda, 1 - (1-c_1)^\lambda, 1 - (1-d_1)^\lambda), (e_1^\lambda, f_1^\lambda, g_1^\lambda h_1^\lambda), (l_1^\lambda, m_1^\lambda, n_1^\lambda p_1^\lambda)\rangle$, $\lambda > 0$

- $\tilde{n}_1^\lambda = \langle (a_1^\lambda, b_1^\lambda, c_1^\lambda, d_1^\lambda), (1 - (1-e_1)^\lambda, 1 - (1-f_1)^\lambda, 1 - (1-g_1)^\lambda, 1 - (1-h_1)^\lambda), (1 - (1-l_1)^\lambda, 1 - (1-m_1)^\lambda, 1 - (1-n_1)^\lambda, 1 - (1-p_1)^\lambda)\rangle$, $\lambda > 0$

## 4. INTERVAL VALUED TRAPEZOIDAL NEUTROSOPHIC SET (IVTrNS)

An interval valued trapezoidal neutrosophic number (IVTrNN) $\widetilde{\widetilde{N}}$ is an interval valued trapezoidal neutrosophic set (IVTrNS) on $X$ is defined by $\widetilde{\widetilde{N}}(x) = \widetilde{\widetilde{N}}^L(x), \widetilde{\widetilde{N}}^U(x)$, where $\widetilde{\widetilde{N}}^L$ and $\widetilde{\widetilde{N}}^U$ are lower and upper trapezoidal neutrosophic sets of $\widetilde{\widetilde{N}}$ such that $\widetilde{\widetilde{N}}^L \subseteq \widetilde{\widetilde{N}}^U$.

$\widetilde{\widetilde{N}}^L = \{\langle x, T^L_{\widetilde{\widetilde{N}}}(x), I^L_{\widetilde{\widetilde{N}}}(x), F^L_{\widetilde{\widetilde{N}}}(x)\rangle: x \in X\}$ where $T^L_{\widetilde{\widetilde{N}}}(x) \subset [0,1]$, $I^L_{\widetilde{\widetilde{N}}}(x) \subset [0,1]$ and $F^L_{\widetilde{\widetilde{N}}}(x) \subset [0,1]$ are trapezoidal neutrosophic fuzzy numbers. It means

$$T^L_{\widetilde{\widetilde{N}}}(x) = \left(t^{L1}_{\widetilde{\widetilde{N}}}(x), t^{L2}_{\widetilde{\widetilde{N}}}(x), t^{L3}_{\widetilde{\widetilde{N}}}(x), t^{L4}_{\widetilde{\widetilde{N}}}(x)\right): X \to [0,1],$$

$$I^L_{\widetilde{\widetilde{N}}}(x) = \left(i^{L1}_{\widetilde{\widetilde{N}}}(x), i^{L2}_{\widetilde{\widetilde{N}}}(x), i^{L3}_{\widetilde{\widetilde{N}}}(x), i^{L4}_{\widetilde{\widetilde{N}}}(x)\right): X \to [0,1],$$

$$\text{and } F^L_{\widetilde{\widetilde{N}}}(x) = \left(f^{L1}_{\widetilde{\widetilde{N}}}(x), f^{L2}_{\widetilde{\widetilde{N}}}(x), f^{L3}_{\widetilde{\widetilde{N}}}(x), f^{L4}_{\widetilde{\widetilde{N}}}(x)\right): X \to [0,1]$$

with the condition $0 \leq t^{L4}_{\widetilde{N}}(x) + i^{L4}_{\widetilde{N}}(x) + f^{L4}_{\widetilde{N}}(x) \leq 3$

Let these three trapezoidal neutrosophic fuzzy numbers are denoted by $T^L_{\widetilde{\widetilde{N}}}(x) = (\underline{a}, \underline{b}, \underline{c}, \underline{d}): X \to [0,1]$, $I^L_{\widetilde{\widetilde{N}}}(x) = (\underline{e}, \underline{f}, \underline{g}, \underline{h}): X \to [0,1]$ and $F^L_{\widetilde{\widetilde{N}}}(x) = (\underline{l}, \underline{m}, \underline{n}, \underline{p}): X \to [0,1]$. Thus $\widetilde{\widetilde{N}}^L = \{\langle(\underline{a}, \underline{b}, \underline{c}, \underline{d}), (\underline{e}, \underline{f}, \underline{g}, \underline{h}), (\underline{l}, \underline{m}, \underline{n}, \underline{p})\rangle: X \to [0,1]$. If $\underline{b} = \underline{c}$, $\underline{f} = \underline{g}$ and $\underline{m} = \underline{n}$, these trapezoidal neutrosophic fuzzy numbers are reduced to triangular neutrosophic numbers.

*Definition:* Let $\widetilde{\widetilde{N}}^L$ be a lower trapezoidal neutrosophic fuzzy number then $T^L_{\widetilde{\widetilde{N}}}(x), I^L_{\widetilde{\widetilde{N}}}(x)$ and $F^L_{\widetilde{\widetilde{N}}}(x)$ can be defined as follows:

$$T^L{}_{\widetilde{\widetilde{N}}}(x) = \begin{cases} \dfrac{x - \underline{a}}{\underline{b} - \underline{a}} T^L{}_{\widetilde{\widetilde{N}}} & \underline{a} \leq x < \underline{b} \\ T^L{}_{\widetilde{\widetilde{N}}} & \underline{b} \leq x \leq \underline{c} \\ \dfrac{\underline{d} - x}{\underline{d} - \underline{c}} T^L{}_{\widetilde{\widetilde{N}}} & \underline{c} < x \leq \underline{d} \\ 0 & \text{otherwise} \end{cases}$$

$$I^L{}_{\widetilde{\widetilde{N}}}(x) = \begin{cases} \dfrac{\underline{f} - x + I^L{}_{\widetilde{\widetilde{N}}}(x - \underline{e})}{\underline{f} - \underline{e}} & \underline{e} \leq x < \underline{f} \\ I^L{}_{\widetilde{\widetilde{N}}} & \underline{f} \leq x \leq \underline{g} \\ \dfrac{x - \underline{g} + I^L{}_{\widetilde{\widetilde{N}}}(\underline{h} - x)}{\underline{h} - \underline{g}} & \underline{g} < x \leq \underline{h} \\ 1 & \text{otherwise} \end{cases}$$

$$F^L{}_{\widetilde{\widetilde{N}}}(x) = \begin{cases} \dfrac{\underline{m} - x + F^L{}_{\widetilde{\widetilde{N}}}(x - \underline{l})}{\underline{m} - \underline{l}} & \underline{l} \leq x < \underline{m} \\ F^L{}_{\widetilde{\widetilde{N}}} & \underline{m} \leq x \leq \underline{n} \\ \dfrac{x - \underline{n} + F^L{}_{\widetilde{\widetilde{N}}}(\underline{p} - x)}{\underline{p} - \underline{n}} & \underline{n} < x \leq \underline{p} \\ 1 & \text{otherwise} \end{cases}$$

$\widetilde{\widetilde{N}}^U = \{\langle x, T^U{}_{\widetilde{\widetilde{N}}}(x), I^U{}_{\widetilde{\widetilde{N}}}(x), F^U{}_{\widetilde{\widetilde{N}}}(x)\rangle : x \in X\}$ where $T^U{}_{\widetilde{\widetilde{N}}}(x) \subset [0,1], I^U{}_{\widetilde{\widetilde{N}}}(x) \subset [0,1]$ and $F^U{}_{\widetilde{\widetilde{N}}}(x) \subset [0,1]$ are trapezoidal neutrosophic fuzzy numbers. It means

$$T^U{}_{\widetilde{\widetilde{N}}}(x) = \left(t^{U1}_{\widetilde{\widetilde{N}}}(x), t^{U2}_{\widetilde{\widetilde{N}}}(x), t^{U3}_{\widetilde{\widetilde{N}}}(x), t^{U4}_{\widetilde{\widetilde{N}}}(x)\right) : X \to [0,1],$$
$$I^U{}_{\widetilde{\widetilde{N}}}(x) = (i^{U1}_{\widetilde{\widetilde{N}}}(x), i^{U2}_{\widetilde{\widetilde{N}}}(x), i^{U3}_{\widetilde{\widetilde{N}}}(x), i^{U4}_{\widetilde{\widetilde{N}}}(x)) : X \to [0,1]$$
$$\text{and } F^U{}_{\widetilde{\widetilde{N}}}(x) = (f^{U1}_{\widetilde{\widetilde{N}}}(x), f^{U2}_{\widetilde{\widetilde{N}}}(x), f^{U3}_{\widetilde{\widetilde{N}}}(x), f^{U4}_{\widetilde{\widetilde{N}}}(x)) : X \to [0,1]$$

with the condition $0 \leq t^{U4}_{\widetilde{N}}(x) + i^{U4}_{\widetilde{N}}(x) + f^{U4}_{\widetilde{N}}(x) \leq 3$.

Let these three trapezoidal neutrosophic fuzzy numbers are denoted by $T^U{}_{\widetilde{\widetilde{N}}}(x) = (\bar{a}, \bar{b}, \bar{c}, \bar{d}) : X \to [0,1]$, $I^L{}_{\widetilde{\widetilde{N}}}(x) = (\bar{e}, \bar{f}, \bar{g}, \bar{h}) : X \to [0,1]$ and $F^L{}_{\widetilde{\widetilde{N}}}(x) = (\bar{l}, \bar{m}, \bar{n}, \bar{p}) : X \to [0,1]$.

Thus $\widetilde{\widetilde{N}}^U = \{\langle(\bar{a}, \bar{b}, \bar{c}, \bar{d}), (\bar{e}, \bar{f}, \bar{g}, \bar{h}), (\bar{l}, \bar{m}, \bar{n}, \bar{p})\rangle : X \to [0,1]$. If $\bar{b} = \bar{c}$, $\bar{f} = \bar{g}$ and $\bar{m} = \bar{n}$, these trapezoidal neutrosophic fuzzy numbers are reduced to triangular neutrosophic numbers.

*Definition:* Let $\widetilde{\widetilde{N}}^U$ be a upper trapezoidal neutrosophic fuzzy number then $T^U{}_{\widetilde{\widetilde{N}}}(x), I^U{}_{\widetilde{\widetilde{N}}}(x)$ and $F^U{}_{\widetilde{\widetilde{N}}}(x)$ can be defined as follows:

$$T^U{}_{\widetilde{\widetilde{N}}}(x) = \begin{cases} \dfrac{x - \bar{a}}{\bar{b} - \bar{a}} & \bar{a} \leq x < \bar{b} \\ T^U{}_{\widetilde{\widetilde{N}}} & \bar{b} \leq x \leq \bar{c} \\ \dfrac{\bar{d} - x}{\bar{d} - \bar{c}} & \bar{c} < x \leq \bar{d} \\ 0 & \text{otherwise} \end{cases}$$

$$I^U{}_{\widetilde{N}}(x) = \begin{cases} \dfrac{\bar{f} - x + I^U{}_{\widetilde{N}}(x - \bar{e})}{\bar{f} - \bar{e}} & \bar{e} \leq x < \bar{f} \\ I^U{}_{\widetilde{N}} & \bar{f} \leq x \leq \bar{g} \\ \dfrac{x - \bar{g} + I^U{}_{\widetilde{N}}(\bar{h} - x)}{\bar{h} - \bar{g}} & \bar{g} < x \leq \bar{h} \\ 1 & \text{otherwise} \end{cases}$$

$$F^U{}_{\widetilde{N}}(x) = \begin{cases} \dfrac{\bar{m} - x + I^U{}_{\widetilde{N}}(x - \bar{l})}{\bar{m} - \bar{l}} & \bar{l} \leq x < \bar{m} \\ F^U{}_{\widetilde{N}} & \bar{m} \leq x \leq \bar{n} \\ \dfrac{x - \bar{n} + I^U{}_{\widetilde{N}}(\bar{p} - x)}{\bar{p} - \bar{n}} & \bar{n} < x \leq \bar{p} \\ 1 & \text{otherwise} \end{cases}$$

Thus an IVTrNN $\tilde{\tilde{n}}$ is denoted by

$$\tilde{\tilde{n}} = \begin{cases} [(\underline{a},\underline{b},\underline{c},\underline{d}:T^L{}_{\tilde{\tilde{n}}}),(\bar{a},\bar{b},\bar{c},\bar{d}:T^U{}_{\tilde{\tilde{n}}})], \\ [(\underline{e},\underline{f},\underline{g},\underline{h}:I^L{}_{\tilde{\tilde{n}}}),(\bar{e},\bar{f},\bar{g},\bar{h}:I^U{}_{\tilde{\tilde{n}}})], \\ [(\underline{l},\underline{m},\underline{n},\underline{p}:F^L{}_{\tilde{\tilde{n}}}),(\bar{l},\bar{m},\bar{n},\bar{p}:F^U{}_{\tilde{\tilde{n}}})] \end{cases}$$

Further it can be written as

$$\tilde{\tilde{n}} = \begin{cases} \langle[(\underline{a},\underline{b},\underline{c},\underline{d}),(\bar{a},\bar{b},\bar{c},\bar{d})]:T_{\tilde{\tilde{n}}}\rangle, \langle[(\underline{e},\underline{f},\underline{g},\underline{h}),(\bar{e},\bar{f},\bar{g},\bar{h})]:I_{\tilde{\tilde{n}}}\rangle, \\ \langle[(\underline{l},\underline{m},\underline{n},\underline{p}),(\bar{l},\bar{m},\bar{n},\bar{p})]:F_{\tilde{\tilde{n}}}\rangle \end{cases}$$

*Definition:* Let $\tilde{\tilde{n}}_1$ and $\tilde{\tilde{n}}_2$ are two IVTrNNs,

$$\tilde{\tilde{n}}_1 = \begin{cases} \langle[(\underline{a_1},\underline{b_1},\underline{c_1},\underline{d_1}),(\overline{a_1},\overline{b_1},\overline{c_1},\overline{d_1})]:T_{\tilde{\tilde{n}}_1}\rangle, \\ \langle[(\underline{e_1},\underline{f_1},\underline{g_1},\underline{h_1}),(\overline{e_1},\overline{f_1},\overline{g_1},\overline{h_1})]:I_{\tilde{\tilde{n}}_1}\rangle, \\ ,\langle[(\underline{l_1},\underline{m_1},\underline{n_1},\underline{p_1}),(\overline{l_1},\overline{m_1},\overline{n_1},\overline{p_1})]:F_{\tilde{\tilde{n}}_1}\rangle \end{cases}$$

$$\tilde{\tilde{n}}_2 = \begin{cases} \langle[(\underline{a_2},\underline{b_2},\underline{c_2},\underline{d_2}),(\overline{a_2},\overline{b_2},\overline{c_2},\overline{d_2})]:T_{\tilde{\tilde{n}}_2}\rangle, \\ ,\langle[(\underline{e_2},\underline{f_2},\underline{g_2},\underline{h_1}),(\overline{e_2},\overline{f_2},\overline{g_2},\overline{h_2})]:I_{\tilde{\tilde{n}}_2}\rangle \\ \langle[(\underline{l_2},\underline{m_2},\underline{n_2},\underline{p_2}),(\overline{l_2},\overline{m_2},\overline{n_2},\overline{p_2})]:F_{\tilde{\tilde{n}}_2}\rangle \end{cases}$$

Based on the work in [20][41][42][43], Operations on IVTrNNs will be as follows:
- $\tilde{\tilde{n}}_1 \oplus \tilde{\tilde{n}}_2 =$

$$= \begin{cases} \left[\begin{pmatrix} \underline{a_1}+\underline{a_2}-\underline{a_1}\underline{a_2}, & \underline{b_1}+\underline{b_2}-\underline{b_1}\underline{b_2}, \\ \underline{c_1}+\underline{c_2}-\underline{c_1}\underline{c_2}, \underline{d_1}+\underline{d_2}-\underline{d_1}\underline{d_2} \end{pmatrix}, \begin{pmatrix} \overline{a_1}+\overline{a_2}-\overline{a_1}\overline{a_2}, & \overline{b_1}+\overline{b_2}-\overline{b_1}\overline{b_2}, \\ \overline{c_1}+\overline{c_2}-\overline{c_1}\overline{c_2}, \overline{d_1}+\overline{d_2}-\overline{d_1}\overline{d_2} \end{pmatrix}\right], \\ [(\underline{e_1e_2},\underline{f_1f_2},\underline{g_1g_2},\underline{h_1h_2}),(\overline{e_1e_2},\overline{f_1f_2},\overline{g_1g_2},\overline{h_1h_2})], \\ [(\underline{l_1l_2},\underline{m_1m_2},\underline{n_1n_2},\underline{p_1p_2}),(\overline{l_1l_2},\overline{m_1m_2},\overline{n_1n_2},\overline{p_1p_2})] \end{cases}$$

- $\tilde{n}_1 \otimes \tilde{n}_2 =$

$$= \left\langle \begin{array}{c} \left[\left(\underline{a_1 a_2}, \underline{b_1 b_2}, \underline{c_1 c_2}, \underline{d_1 d_2}\right), \left(\overline{a_1 a_2}, \overline{b_1 b_2}, \overline{c_1 c_2}, \overline{d_1 d_2}\right)\right], \\ \left[\begin{pmatrix} \underline{e_1} + \underline{e_2} - \underline{e_1 e_2}, \underline{f_1} + \underline{f_2} - \underline{f_1 f_2}, \\ \underline{g_1} + \underline{g_2} - \underline{g_1 g_2}, \underline{h_1} + \underline{h_2} - \underline{h_1 h_2} \end{pmatrix}, \\ \begin{pmatrix} \overline{e_1} + \overline{e_2} - \overline{e_1 e_2}, \overline{f_1} + \overline{f_2} - \overline{f_1 f_2}, \\ \overline{g_1} + \overline{g_2} - \overline{g_1 g_2}, \overline{h_1} + \overline{h_2} - \overline{h_1 h_2} \end{pmatrix}\right], \\ \left[\begin{pmatrix} \underline{l_1} + \underline{l_2} - \underline{l_1 l_2}, \underline{m_1} + \underline{m_2} - \underline{m_1 m_2}, \\ \underline{n_1} + \underline{n_2} - \underline{n_1 n_2}, \underline{p_1} + \underline{p_2} - \underline{p_1 p_2} \end{pmatrix}, \\ \begin{pmatrix} \overline{l_1} + \overline{l_2} - \overline{l_1 l_2}, \overline{m_1} + \overline{m_2} - \overline{m_1 m_2}, \\ \overline{n_1} + \overline{n_2} - \overline{n_1 n_2}, \overline{p_1} + \overline{p_2} - \overline{p_1 p_2} \end{pmatrix}\right] \end{array} \right\rangle$$

- $\lambda \tilde{n}_1 =$

$$= \left\langle \begin{array}{c} \left[\begin{pmatrix} 1 - \left(1 - \underline{a_1}\right)^\lambda, 1 - \left(1 - \underline{b_1}\right)^\lambda, \\ 1 - \left(1 - \underline{c_1}\right)^\lambda, 1 - \left(1 - \underline{d_1}\right)^\lambda \end{pmatrix}, \\ \begin{pmatrix} 1 - (1 - \overline{a_1})^\lambda, 1 - (1 - \overline{b_1})^\lambda, \\ 1 - (1 - \overline{c_1})^\lambda, 1 - (1 - \overline{d_1})^\lambda \end{pmatrix}\right], \\ \left[\left(\underline{e_1}^\lambda, \underline{f_1}^\lambda, \underline{g_1}^\lambda, \underline{h_1}^\lambda\right), \left(\overline{e_1}^\lambda, \overline{f_1}^\lambda, \overline{g_1}^\lambda, \overline{h_1}^\lambda\right)\right], \\ \left[\left(\underline{l_1}^\lambda, \underline{m_1}^\lambda, \underline{n_1}^\lambda, \underline{p_1}^\lambda\right), \left(\overline{l_1}^\lambda, \overline{m_1}^\lambda, \overline{n_1}^\lambda, \overline{p_1}^\lambda\right)\right] \end{array} \right\rangle, \quad \lambda > 0$$

- $\tilde{n}_1^\lambda =$

$$= \left\langle \begin{array}{c} \left[\left(\underline{a_1}^\lambda, \underline{b_1}^\lambda, \underline{c_1}^\lambda, \underline{d_1}^\lambda\right), \left(\overline{a_1}^\lambda, \overline{b_1}^\lambda, \overline{c_1}^\lambda, \overline{d_1}^\lambda\right)\right], \\ \left[\begin{pmatrix} 1 - \left(1 - \underline{e_1}\right)^\lambda, 1 - \left(1 - \underline{f_1}\right)^\lambda, \\ 1 - \left(1 - \underline{g_1}\right)^\lambda, 1 - \left(1 - \underline{h_1}\right)^\lambda \end{pmatrix}, \\ \begin{pmatrix} 1 - (1 - \overline{e_1})^\lambda, 1 - (1 - \overline{f_1})^\lambda, \\ 1 - (1 - \overline{g_1})^\lambda, 1 - (1 - \overline{h_1})^\lambda \end{pmatrix}\right], \\ \left[\begin{pmatrix} 1 - \left(1 - \underline{l_1}\right)^\lambda, 1 - \left(1 - \underline{m_1}\right)^\lambda, \\ 1 - \left(1 - \underline{n_1}\right)^\lambda, 1 - \left(1 - \underline{p_1}\right)^\lambda \end{pmatrix}, \\ \begin{pmatrix} 1 - (1 - \overline{l_1})^\lambda, 1 - (1 - \overline{m_1})^\lambda, \\ 1 - (1 - \overline{n_1})^\lambda, 1 - (1 - \overline{p_1})^\lambda \end{pmatrix}\right] \end{array} \right\rangle, \lambda > 0$$

## 5. SCORE AND ACCURACY FUNCTIONS

Based on the score and accuracy functions for a trapezoidal neutrosophic number [20][44], the score function for IVTrNN $\tilde{n}$ are defined as follows

$$S(\tilde{n}) = \frac{1}{6}\left[4 + \frac{\underline{a} + \underline{b} + \underline{c} + \underline{d}}{4} + \frac{\bar{a} + \bar{b} + \bar{c} + \bar{d}}{4} - \frac{\underline{e} + \underline{f} + \underline{g} + \underline{h}}{4} - \frac{\bar{e} + \bar{f} + \bar{g} + \bar{h}}{4} - \frac{\underline{l} + \underline{m} + \underline{n} + \underline{p}}{4} - \frac{\bar{l} + \bar{m} + \bar{n} + \bar{p}}{4}\right], \quad S(\tilde{n}) \in [0,1] \tag{10}$$

where larger the value of $S(\tilde{n})$, higher the IVTrNN $\tilde{n}$.

Especially if $S(\tilde{n}) = 1$, then $\tilde{n} = \langle[(1,1,1,1),(1,1,1,1)],[(0,0,0,0),(0,0,0,0)],[(0,0,0,0),(0,0,0,0)]\rangle$, which is the largest IVTrNN;

if $S(\tilde{n}) = 0$, then $\tilde{n} = \langle[(0,0,0,0),(0,0,0,0)],[(1,1,1,1),(1,1,1,1)],[(1,1,1,1),(1,1,1,1)]\rangle$, which is the smallest IVTrNN.

When $\underline{b} = \underline{c}, \underline{f} = \underline{g}, \underline{m} = \underline{n}, \bar{b} = \bar{c}, \bar{f} = \bar{g}$ and $\bar{m} = \bar{n}$ hold in an IVTrNN $\tilde{n}$; score function reduces to the following:

$$S(\tilde{n}) = \frac{1}{6}\left[4 + \frac{\underline{a} + 2\underline{b} + \underline{d}}{4} + \frac{\bar{a} + 2\bar{b} + \bar{d}}{4} - \frac{\underline{e} + 2\underline{f} + \underline{h}}{4} - \frac{\bar{e} + 2\bar{f} + \bar{h}}{4} - \frac{\underline{l} + 2\underline{m} + \underline{p}}{4} - \frac{\bar{l} + 2\bar{m} + \bar{p}}{4}\right],$$

$S(\tilde{n}) \in [0,1]$  (11)

Now, the accuracy function for IVTrNN $\tilde{n}$ can be defined as

$$H(\tilde{n}) = \frac{1}{2}\left[\frac{\underline{a} + \underline{b} + \underline{c} + \underline{d}}{4} + \frac{\bar{a} + \bar{b} + \bar{c} + \bar{d}}{4} - \frac{\underline{l} + \underline{m} + \underline{n} + \underline{p}}{4} - \frac{\bar{l} + \bar{m} + \bar{n} + \bar{p}}{4}\right], H(\tilde{n}) \in [-1,1] \quad (12)$$

where larger the value of $H(\tilde{n})$, the higher the degree of accuracy of IVTrNN $\tilde{n}$. When $\underline{b} = \underline{c}, \underline{f} = \underline{g}, \underline{m} = \underline{n}, \bar{b} = \bar{c}, \bar{f} = \bar{g}$ and $\bar{m} = \bar{n}$ hold in an IVTrNN $\tilde{n}$; it reduces to the following:

$$H(\tilde{n}) = \frac{1}{2}\left[\frac{\underline{a} + 2\underline{b} + \underline{d}}{4} + \frac{\bar{a} + 2\bar{b} + \bar{d}}{4} - \frac{\underline{l} + 2\underline{m} + \underline{p}}{4} - \frac{\bar{l} + 2\bar{m} + \bar{p}}{4}\right], H(\tilde{n}) \in [-1,1] \quad (13)$$

*Definition:*

Let $\tilde{n}_1$ and $\tilde{n}_2$ are two IVTrNNs:

then $S(\tilde{n}_1), S(\tilde{n}_2), H(\tilde{n}_1)$ and $H(\tilde{n}_2)$ are the scores and accuracy degrees of $\tilde{n}_1$ and $\tilde{n}_2$ respectively.
- If $S(\tilde{n}_1) > S(\tilde{n}_2)$, then $\tilde{n}_1 > \tilde{n}_2$;
- If $S(\tilde{n}_1) = S(\tilde{n}_2)$, and
  (a) If $H(\tilde{n}_1) = H(\tilde{n}_2)$, then $\tilde{n}_1 = \tilde{n}_2$;
  (b) If $H(\tilde{n}_1) > H(\tilde{n}_2)$, then $\tilde{n}_1 > \tilde{n}_2$.

## 6. INTERVAL VALUED TRAPEZOIDAL NEUTROSOPHIC WEIGHTED ARITHMETIC AVERAGING OPERATOR (IVTrNWAA)

Based on the operations proposed of an IVTrNN number, we propose the following aggregation operator:
Let

$$\tilde{n}_j = \begin{cases} \langle[(\underline{a}_j,\underline{b}_j,\underline{c}_j,\underline{d}_j),(\bar{a}_j,\bar{b}_j,\bar{c}_j,\bar{d}_j)]: T_{\tilde{n}_j}\rangle, \\ \langle[(\underline{e}_j,\underline{f}_j,\underline{g}_j,\underline{h}_j),(\bar{e}_j,\bar{f}_j,\bar{g}_j,\bar{h}_j)]: I_{\tilde{n}_j}\rangle, \\ \langle[(\underline{l}_j,\underline{m}_j,\underline{n}_j,\underline{p}_j),(\bar{l}_j,\bar{m}_j,\bar{n}_j,\bar{p}_j)]: F_{\tilde{n}_j}\rangle \end{cases},$$

$(j = 1,2,3,\cdots,n)$

be a group of IVTrNNs. The IVTrNNWA is represented as:

$$IVTrNWAA(\tilde{n}_1,\tilde{n}_2,\ldots,\tilde{n}_n) = w_1\tilde{n}_1 \oplus w_2\tilde{n}_2 \oplus w_3\tilde{n}_3 \oplus \ldots \oplus w_n\tilde{n}_n = \bigoplus_{j=1}^{n}(w_j\tilde{n}_j) \quad (14)$$

where $w_j (j = 1,2,3,\cdots,n)$ is the weight of the interval valued trapezoidal neutrosphic number $\tilde{n}_j (j = 1,2,3,\cdots,n)$ with $w_j \in [0,1]$ and $\sum_{j=1}^{n} w_j = 1$. Based on the operational rules of IVTrNN, we can derive the *IVTrNWAA* for two interval valued trapezoidal neutrosophic numbers:

$$IVTrNWAA(\tilde{n}_1, \tilde{n}_2) = w_1\tilde{n}_1 \oplus w_2\tilde{n}_2 = \left\langle \begin{bmatrix} \begin{pmatrix} 1-\left(1-\underline{a_1}\right)^{w_1} + 1-\left(1-\underline{a_2}\right)^{w_2} \\ -\left(1-\left(1-\underline{a_1}\right)^{w_1}\right)\left(1-\left(1-\underline{a_2}\right)^{w_2}\right), \\ 1-\left(1-\underline{b_1}\right)^{w_1} + 1-\left(1-\underline{b_2}\right)^{w_2} \\ -\left(1-\left(1-\underline{b_1}\right)^{w_1}\right)\left(1-\left(1-\underline{b_2}\right)^{w_2}\right), \\ 1-\left(1-\underline{c_1}\right)^{w_1} + 1-\left(1-\underline{c_2}\right)^{w_2} \\ -\left(1-\left(1-\underline{c_1}\right)^{w_1}\right)\left(1-\left(1-\underline{c_2}\right)^{w_2}\right), \\ 1-\left(1-\underline{d_1}\right)^{w_1} + 1-\left(1-\underline{d_2}\right)^{w_2} \\ -\left(1-\left(1-\underline{d_1}\right)^{w_1}\right)\left(1-\left(1-\underline{d_2}\right)^{w_2}\right) \end{pmatrix}, \\ \begin{pmatrix} 1-(1-\overline{a_1})^{w_1} + 1-(1-\overline{a_2})^{w_2} \\ -(1-(1-\overline{a_1})^{w_1})(1-(1-\overline{a_2})^{w_2}), \\ 1-(1-\overline{b_1})^{w_1} + 1-(1-\overline{b_2})^{w_2} \\ -(1-(1-\overline{b_1})^{w_1})(1-(1-\overline{b_2})^{w_2}), \\ 1-(1-\overline{c_1})^{w_1} + 1-(1-\overline{c_2})^{w_2} \\ -(1-(1-\overline{c_1})^{w_1})(1-(1-\overline{c_2})^{w_2}), \\ 1-\left(1-\overline{d_1}\right)^{w_1} + 1-\left(1-\overline{d_2}\right)^{w_2} - \\ \left(1-\left(1-\overline{d_1}\right)^{w_1}\right)\left(1-\left(1-\overline{d_2}\right)^{w_2}\right) \end{pmatrix} \end{bmatrix}, \\ \begin{bmatrix} \left(\underline{e_1}^{w_1}\underline{e_2}^{w_2}, \underline{f_1}^{w_1}\underline{f_2}^{w_2}, \underline{g_1}^{w_1}\underline{g_2}^{w_2}, \underline{h_1}^{w_1}\underline{h_2}^{w_2}\right), \\ \left(\overline{e_1}^{w_1}\overline{e_2}^{w_2}, \overline{f_1}^{w_1}\overline{f_2}^{w_2}, \overline{g_1}^{w_1}\overline{g_2}^{w_2}, \overline{h_1}^{w_1}\overline{h_2}^{w_2}\right) \end{bmatrix}, \\ \begin{bmatrix} \left(\underline{l_1}^{w_1}\underline{l_2}^{w_2}, \underline{m_1}^{w_1}\underline{m_2}^{w_2}, \underline{n_1}^{w_1}\underline{n_2}^{w_2}, \underline{p_1}^{w_1}\underline{p_2}^{w_2}\right), \\ \left(\overline{l_1}^{w_1}\overline{l_1}^{w_2}, \overline{m_1}^{w_1}\overline{m_2}^{w_2}, \overline{n_1}^{w_1}\overline{n_2}^{w_2}, \overline{p_1}^{w_1}\overline{p_2}^{w_2}\right) \end{bmatrix} \end{bmatrix} \right\rangle$$

$$= \left\langle \begin{bmatrix} \begin{pmatrix} 1-(1-\underline{a_1})^{w_1}(1-\underline{a_2})^{w_2}, 1-(1-\underline{b_1})^{w_1}(1-\underline{b_2})^{w_2}, \\ 1-(1-\underline{c_1})^{w_1}(1-\underline{c_2})^{w_2}, 1-(1-\underline{d_1})^{w_1}(1-\underline{d_2})^{w_2} \end{pmatrix}, \\ \begin{pmatrix} 1-(1-\overline{a_1})^{w_1}(1-\overline{a_2})^{w_2}, 1-(1-\overline{b_1})^{w_1}(1-\overline{b_2})^{w_2}, \\ 1-(1-\overline{c_1})^{w_1}(1-\overline{c_2})^{w_2}, 1-(1-\overline{d_1})^{w_1}(1-\overline{d_2})^{w_2} \end{pmatrix} \end{bmatrix}, \\ \begin{bmatrix} \left(\prod_{j=1}^{2}\underline{e_j}^{w_j}, \prod_{j=1}^{2}\underline{f_j}^{w_j}, \prod_{j=1}^{2}\underline{g_j}^{w_j}, \prod_{j=1}^{2}\underline{h_j}^{w_j}\right), \\ \left(\prod_{j=1}^{2}\overline{e_j}^{w_j}, \prod_{j=1}^{2}\overline{f_j}^{w_j}, \prod_{j=1}^{2}\overline{g_j}^{w_j}, \prod_{j=1}^{2}\overline{h_j}^{w_j}\right) \end{bmatrix}, \\ \begin{bmatrix} \left(\prod_{j=1}^{2}\underline{l_j}^{w_j}, \prod_{j=1}^{2}\underline{m_j}^{w_j}, \prod_{j=1}^{2}\underline{n_j}^{w_j}, \prod_{j=1}^{2}\underline{p_j}^{w_j}\right), \\ \left(\prod_{j=1}^{2}\overline{l_j}^{w_j}, \prod_{j=1}^{2}\overline{m_j}^{w_j}, \prod_{j=1}^{2}\overline{n_j}^{w_j}, \prod_{j=1}^{2}\overline{p_j}^{w_j}\right) \end{bmatrix} \end{bmatrix} \right\rangle$$

Similarly for $n$ number of interval valued trapezoidal neutrosophic numbers, it can be generalized as follows:

$$IVTrNWAA(\tilde{\tilde{n}}_1, \tilde{\tilde{n}}_2, \ldots, \tilde{\tilde{n}}_n) = w_1\tilde{\tilde{n}}_1 \oplus w_2\tilde{\tilde{n}}_2 \oplus \tilde{\tilde{n}}_3 \oplus \ldots \oplus w_n\tilde{\tilde{n}}_n = \bigoplus_{j=1}^{n}(w_j\tilde{\tilde{n}}_j)$$

$$= \left\langle \begin{matrix} \left[ \begin{pmatrix} 1 - \prod_{j=1}^{n}(1-\underline{a_j})^{w_j}, 1 - \prod_{j=1}^{n}(1-\underline{b_j})^{w_j}, \\ 1 - \prod_{j=1}^{n}(1-\underline{c_j})^{w_j}, 1 - \prod_{j=1}^{n}(1-\underline{d_j})^{w_j} \end{pmatrix}, \\ \begin{pmatrix} 1 - \prod_{j=1}^{n}(1-\overline{a_j})^{w_j}, 1 - \prod_{j=1}^{n}(1-\overline{b_j})^{w_j}, \\ 1 - \prod_{j=1}^{n}(1-\overline{c_j})^{w_j}, 1 - \prod_{j=1}^{n}(1-\overline{d_j})^{w_j} \end{pmatrix} \right], \\ \left[ \begin{pmatrix} \prod_{j=1}^{n}\underline{e_j}^{w_j} \prod_{j=1}^{n}\underline{f_j}^{w_j} \prod_{j=1}^{n}\underline{g_j}^{w_j} \prod_{j=1}^{n}\underline{h_j}^{w_j} \end{pmatrix}, \\ \begin{pmatrix} \prod_{j=1}^{n}\overline{e_j}^{w_j} \prod_{j=1}^{n}\overline{f_j}^{w_j} \prod_{j=1}^{n}\overline{g_j}^{w_j} \prod_{j=1}^{n}\overline{h_j}^{w_j} \end{pmatrix} \right], \\ \left[ \begin{pmatrix} \prod_{j=1}^{n}\underline{l_j}^{w_j} \prod_{j=1}^{n}\underline{m_j}^{w_j} \prod_{j=1}^{n}\underline{n_j}^{w_j} \prod_{j=1}^{n}\underline{p_j}^{w_j} \end{pmatrix}, \\ \begin{pmatrix} \prod_{j=1}^{n}\overline{l_j}^{w_j} \prod_{j=1}^{n}\overline{m_j}^{w_j} \prod_{j=1}^{n}\overline{n_j}^{w_j} \prod_{j=1}^{n}\overline{p_j}^{w_j} \end{pmatrix} \right] \end{matrix} \right\rangle \quad (15)$$

## 7. MULTI ATTRIBUTE DECISION MAKING USING IVTrNWAA

We have proposed an approach to resolve MADM problems with trapezoidal information under interval valued neutrosophic environment. Let $A$ as a set of alternatives $A = (A_1, A_2, A_3, \ldots, A_n)$ which satisfies a set of attributes $C = (C_1, C_2, C_3, \ldots, C_n)$. The experts evaluates each alternatives based on the attributes represented in the form of interval valued trapezoidal neutrosophic numbers. Therefore, we can get an interval valued trapezoidal neutrosophic decision matrix

$$D = \left(\tilde{\tilde{d}}_{ij}\right)_{m \times n} = \begin{pmatrix} \langle [(\underline{a_{ij}}, \underline{b_{ij}}, \underline{c_{ij}}, \underline{d_{ij}}), (\overline{a_{ij}}, \overline{b_{ij}}, \overline{c_{ij}}, \overline{d_{ij}})] \rangle, \\ \langle [(\underline{e_{ij}}, \underline{f_{ij}}, \underline{g_{ij}}, \underline{h_{ij}}), (\overline{e_{ij}}, \overline{f_{ij}}, \overline{g_{ij}}, \overline{h_{ij}})] \rangle, \\ \langle [(\underline{l_{ij}}, \underline{m_{ij}}, \underline{n_{ij}}, \underline{p_{ij}}), (\overline{l_{ij}}, \overline{m_{ij}}, \overline{n_{ij}}, \overline{p_{ij}})] \rangle \end{pmatrix}_{m \times n}$$

where

$\left(\underline{a_{ij}}, \underline{b_{ij}}, \underline{c_{ij}}, \underline{d_{ij}}\right) \subset [0,1]$ and $(\overline{a_{ij}}, \overline{b_{ij}}, \overline{c_{ij}}, \overline{d_{ij}}) \subset [0,1]$ refers to lower and upper degree of satisfaction of the attribute $C_j$ by alternative $A_i$ respectively;

$\left(\underline{e_{ij}}, \underline{f_{ij}}, \underline{g_{ij}}, \underline{h_{ij}}\right) \subset [0,1]$ and $(\overline{e_{ij}}, \overline{f_{ij}}, \overline{g_{ij}}, \overline{h_{ij}}) \subset [0,1]$ refers to lower and upper degree of uncertainty of attribute $C_j$ with respect to alternative $A_i$ respectively;

$\left(\underline{l_{ij}}, \underline{m_{ij}}, \underline{n_{ij}}, \underline{p_{ij}}\right) \subset [0,1]$ and $(\overline{l_{ij}}, \overline{m_{ij}}, \overline{n_{ij}}, \overline{p_{ij}}) \subset [0,1]$ refers to lower and upper degree of dissatisfaction of attribute $C_j$ with respect to alternative $A_i$ respectively with the conditions $0 \leq \underline{d_{ij}} + \underline{h_{ij}} + \underline{p_{ij}} \leq 3$ and $0 \leq \overline{d_{ij}} + \overline{h_{ij}} + \overline{p_{ij}} \leq 3$ for $i = 1, 2, 3, \ldots, m$ and $j = 1, 2, 3, \ldots, n$.

*7.1 Pseudocode for Multiple Attribute Decision-Making Problems using IVTrNWAA Operator*

*Step1:* Implement the IVTrNWAA operator

$$\tilde{\tilde{d}}_i = \begin{pmatrix} \langle [(a_i, b_i, c_i, d_i), (\overline{a}_\iota, \overline{b}_\iota, \overline{c}_\iota, \overline{d}_\iota)] \rangle, \\ \langle [(e_i, f_i, g_i, h_i), (\overline{e}_\iota, \overline{f}_\iota, \overline{g}_\iota, \overline{h}_\iota)] \rangle \\ , \langle [(l_i, m_i, n_i, p_i), (\overline{l}_\iota, \overline{m}_\iota, \overline{n}_\iota, \overline{p}_\iota)] \rangle \end{pmatrix} = IVTrNWAA(\tilde{\tilde{d}}_{i1}, \tilde{\tilde{d}}_{i2}, \tilde{\tilde{d}}_{i3}, \dots, \tilde{\tilde{d}}_{in})$$

to get the combined IVTrNNs in the form of $\tilde{\tilde{d}}_i (i = 1,2,3,\dots,m)$ for each alternative $A_i (i = 1,2,3,\dots,m)$.

*Step2:* Calculate the score $S(\tilde{\tilde{d}}_i)(i = 1,2,3,\dots,m)$ and $S(\tilde{\tilde{d}}_j)(i = 1,2,3,\dots,m)$ of the combined IVTrNNs of $\tilde{\tilde{d}}_i (i = 1,2,3,\dots,m)$ and $\tilde{\tilde{d}}_j (j = 1,2,3,\dots,m)$ to rank the alternatives $A_i(i = 1,2,3,\dots,m)$ and $A_j(j = 1,2,3,\dots,m)$. If there is no difference between $S(\tilde{\tilde{d}}_i)$ and $S(\tilde{\tilde{d}}_j)$, then calculate the accuracy degrees $H(\tilde{\tilde{d}}_i)$ and $H(\tilde{\tilde{d}}_j)$ of the combined interval valued trapezoidal neutrosophic numbers respectively. Then rank the alternatives $A_i(i = 1,2,3,\dots,m)$ and $A_j(j = 1,2,3,\dots,m)$ on the basis of accuracy degrees $H(\tilde{\tilde{d}}_i)$ and $H(\tilde{\tilde{d}}_j)$.

*Step3:* Rank all the alternatives of $A_i(i = 1,2,3,\dots,m)$ on the basis of $S(\tilde{\tilde{d}}_i)$ and $H(\tilde{\tilde{d}}_i)$, $i = 1,2,3,\dots,m$ and select the best one.

*Step4:* End

## 8. NUMERICAL EXAMPLE

In this section, proposed methodology based on the Interval Valued Neutrosophic Weighted Arithmetic Averaging Operator(IVTrNWAA) is applied for NFR prioritization under an interval valued trapezoidal neutrosophic environment.

The software system must check the authenticity of the users before it allows access to the data. There are various ways to authenticate the users such as password, two factor authentication, finger print recognition, iris recognition, etc. Thus with plenty of solutions available in the market whether it is content based authentication or context based authentication, it is important that organizations must carefully choose an authentication mechanism which have wider user accessibility and acceptability in mind; provide robustness; shall be fast enough to respond with respect to local or remote access; and reliable and resistant to attack. There should be nonfunctional requirements associated with these functionalities. Following is the NFRs bucket:

a) Usability: refers to the user acceptance of authentication system.
b) Performance: refers to the time needed to handle the user's authentication. For example: Authentication system must respond on an average within 3 seconds to local user requests and within 4 seconds to remote user requests
c) Reliability: Reliable authentication is the basis for protecting the valuable data from theft, misuse, and fraud
d) Robustness: must be robust against uncertainty and attacks
e) Security: Invalid user must not be able to breach the protected resource

We cannot achieve these NFRs to the same extent because fulfilment of one requirement may compromise the level of satisfaction of another requirement. The users may not find the multi-level authentication system usable as people tend to have easiness in getting access to the desired information. Thus the use of multi-level authentication satisfies the security non-functional requirement but compromises the usability requirement. Different stakeholders will have different level of concern for different non-functional requirements. The end user may give more priority to Usability followed by performance, security, reliability and robustness.

Here, we are using eight different types of alternatives for implementing the authentication system i.e. (1) Password (PW) (2) Two factor authentication (TF) (3) Captcha Test (CT) (4) Fingerprint Recognition (FR) (5) Iris Recognition (IR) (6) Smart Card (SM) (7) Memory Cards (MM) (8) Cryptographic keys (CK).

Thus we have a set of eight alternatives A = (PW, TF, CT, FR, IR, SM, MM, CK) and the expert must choose among one of these alternative according to five criteria: (1) USF (Usability); (2) PER (Performance); (3) REL (Reliability); (4) RBS (Robustness) and (5) SEC (SECURITY). For some experts, performance may be the important factor to decide the authentication mechanism whereas for some others, security and reliability of an authentication system matters. Assuming weight vector of expert for these five attributes is $W = (0.2, 0.25, 0.25, 0.1, 0.2)^T$.

During the evaluation process, the experts devised four different evaluation criteria for each alternative in terms of trapezoidal neutrosophic numbers and these values are represented by linguistic variables in Table I as follows:

TABLE I: LINGUISTIC VARIABLES OF EVALUATION CRITERIA DEVELOPED BY AN EXPERT(TRAPEZOIDAL NEUTROSOPHIC NUMBER)

| Linguistic Meaning | Trapezoidal Neutrosophic Number |
|---|---|
| Very Low | ((0.0,0.1,0.1,0.2), (0.1,0.1,0.1,0.1), (0.6,0.7,0.8,0.9)) |
| Low | ((0.2,0.3,0.4,0.5), (0.0,0.1,0.2,0.3), (0,0.1,0.2,0.2)) |
| High | ((0.4,0.5,0.6,0.7), (0.0,0.1,0.2,0.3), (0.1,0.1, 0.1, 0.1)) |
| Very High | ((0.7,0.7,0.7,0.7), (0.0,0.1,0.2,0.3), (0.1,0.1, 0.1, 0.1)) |

Then, the expert was asked to give the interval decision matrix and it is be represented in Table II as follows:

TABLE II: INTERVAL DECISION MATRIX

|    | USF | PER | REL | RBS | SEC |
|----|-----|-----|-----|-----|-----|
| PW | [Low, High] | [Very Low, Very High] | [Low , Very High] | [Low, High] | (High, Very High) |
| TF | [Very Low, High] | [Very Low, Very High] | [High , Very High] | [Low, Very High] | [Low, High] |
| CT | [Very Low, Very High] | [High, Very High] | [Very Low, Very High] | [Very Low, Very High] | [Low, High] |
| FR | [High, Very High] | [Low, Very High] | [Very Low, High] | [Low, High] | [High, Very High] |
| IR | [Low, Very High] | [High, Very High] | [Low, High] | [Low, Very High] | [High, Very High] |
| SM | [Low, Very High] | [Low, Very High] | [High, Very High] | [High, Very High] | [Very Low, High] |
| MM | [Very Low, Very High] | [Very Low, High] | [Very Low, Very High] | [Low, High] | [High, Very High] |
| CK | [Very Low, Very High] | [High, Very High] | [High, Very High] | [Very Low, High] | [Low, High] |

Further decision matrix is converted in terms of an interval valued trapezoidal neutrosophic numbers, as shown in Table III.

decide the authentication mechanism whereas for some others, security and reliability of an authentication system matters. Assuming weight vector of expert for these five attributes is $W = (0.2, 0.25, 0.25, 0.1, 0.2)^T$.

During the evaluation process, the experts devised four different evaluation criteria for each alternative in terms of trapezoidal neutrosophic numbers and these values are represented by linguistic variables in Table I as follows:

TABLE I: LINGUISTIC VARIABLES OF EVALUATION CRITERIA DEVELOPED BY AN EXPERT(TRAPEZOIDAL NEUTROSOPHIC NUMBER)

| Linguistic Meaning | Trapezoidal Neutrosophic Number |
|---|---|
| Very Low | ((0.0,0.1,0.1,0.2), (0.1,0.1,0.1,0.1), (0.6,0.7,0.8,0.9)) |
| Low | ((0.2,0.3,0.4,0.5), (0.0,0.1,0.2,0.3), (0,0.1,0.2,0.2)) |
| High | ((0.4,0.5,0.6,0.7), (0.0,0.1,0.2,0.3), (0.1,0.1, 0.1, 0.1)) |
| Very High | ((0.7,0.7,0.7,0.7), (0.0,0.1,0.2,0.3), (0.1,0.1, 0.1, 0.1)) |

Then, the expert was asked to give the interval decision matrix and it is be represented in Table II as follows:

TABLE II: INTERVAL DECISION MATRIX

|    | USF | PER | REL | RBS | SEC |
|----|-----|-----|-----|-----|-----|
| PW | [Low, High] | [Very Low, Very High] | [Low , Very High] | [Low, High] | (High, Very High) |
| TF | [Very Low, High] | [Very Low, Very High] | [High , Very High] | [Low, Very High] | [Low, High] |
| CT | [Very Low, Very High] | [High, Very High] | [Very Low, Very High] | [Very Low, Very High] | [Low, High] |
| FR | [High, Very High] | [Low, Very High] | [Very Low, High] | [Low, High] | [High, Very High] |
| IR | [Low, Very High] | [High, Very High] | [Low, High] | [Low, Very High] | [High, Very High] |
| SM | [Low, Very High] | [Low, Very High] | [High, Very High] | [High, Very High] | [Very Low, High] |
| MM | [Very Low, Very High] | [Very Low, High] | [Very Low, Very High] | [Low, High] | [High, Very High] |
| CK | [Very Low, Very High] | [High, Very High] | [High, Very High] | [Very Low, High] | [Low, High] |

Further decision matrix is converted in terms of an interval valued trapezoidal neutrosophic numbers, as shown in Table III.

TABLE III: THE INTERVAL VALUED TRAPEZOIDAL NEUTROSOPHIC DECISION MATRIX ABOUT EIGHT ALTERNATIVES

|    | USF | PER | REL | RBS | SEC |
|----|-----|-----|-----|-----|-----|
| PW | [((0.2,0.3,0.4,0.5), (0,0.1,0.2,0.3), (0,0.1,0.2,0.2)), ((0.4,0.5,0.6,0.7), (0.0,0.1,0.2,0.3), (0.1,0.1,0.1,0.1))] | [((0.0,0.1,0.1,0.2), (0.1,0.1,0.1,0.1), (0.6,0.7,0.8,0.9)), ((0.7, 0.7,0.7,0.7), (0.0,0.1,0.2,0.3), (0.1,0.1,0.1,0.1))] | [((0.2,0.3,0.4,0.5), (0,0.1,0.2,0.3), (0,0.1,0.2,0.2)), ((0.7, 0.7,0.7,0.7), (0.0,0.1,0.2,0.3), (0.1,0.1,0.1,0.1))] | [((0.2,0.3,0.4,0.5), (0,0.1,0.2,0.3), (0,0.1,0.2,0.2)), ((0.4,0.5,0.6,0.7), (0.0,0.1,0.2,0.3), (0.1,0.1,0.1,0.1))] | ((0.4,0.5,0.6,0.7), (0.0,0.1,0.2,0.3), (0.1,0.1,0.1,0.1)), ((0.7, 0.7,0.7,0.7), (0.0,0.1,0.2,0.3), (0.1,0.1,0.1,0.1))] |
| TF | [((0.0,0.1,0.1,0.2), (0.1,0.1,0.1,0.1), (0.6,0.7,0.8,0.9)), ((0.4,0.5,0.6,0.7), (0.0,0.1,0.2,0.3), (0.1,0.1,0.1,0.1))] | [((0.0,0.1,0.1,0.2), (0.1,0.1,0.1,0.1), (0.6,0.7,0.8,0.9)), ((0.7, 0.7,0.7,0.7), (0.0,0.1,0.2,0.3), (0.1,0.1,0.1,0.1))] | [((0.4,0.5,0.6,0.7), (0.0,0.1,0.2,0.3), (0.1,0.1,0.1,0.1)), ((0.7, 0.7,0.7,0.7), (0.0,0.1,0.2,0.3), (0.1,0.1,0.1,0.1))] | [((0.2,0.3,0.4,0.5), (0,0.1,0.2,0.3), (0,0.1,0.2,0.2)), ((0.7, 0.7,0.7,0.7), (0.0,0.1,0.2,0.3), (0.1,0.1,0.1,0.1))] | [((0.2,0.3,0.4,0.5), (0,0.1,0.2,0.3), (0,0.1,0.2,0.2)), ((0.4,0.5,0.6,0.7), (0.0,0.1,0.2,0.3), (0.1,0.1,0.1,0.1))] |
| CT | [((0.0,0.1,0.1,0.2), (0.1,0.1,0.1,0.1), (0.6,0.7,0.8,0.9)), ((0.7, 0.7,0.7,0.7), (0.0,0.1,0.2,0.3), (0.1,0.1,0.1,0.1))] | [((0.4,0.5,0.6,0.7), (0.0,0.1,0.2,0.3), (0.1,0.1,0.1,0.1)), ((0.7, 0.7,0.7,0.7), (0.0,0.1,0.2,0.3), (0.1,0.1,0.1,0.1))] | [((0.0,0.1,0.1,0.2), (0.1,0.1,0.1,0.1), (0.6,0.7,0.8,0.9)), ((0.7, 0.7,0.7,0.7), (0.0,0.1,0.2,0.3), (0.1,0.1,0.1,0.1))] | [((0.0,0.1,0.1,0.2), (0.1,0.1,0.1,0.1), (0.6,0.7,0.8,0.9)), ((0.7, 0.7,0.7,0.7), (0.0,0.1,0.2,0.3), (0.1,0.1,0.1,0.1))] | [((0.2,0.3,0.4,0.5), (0,0.1,0.2,0.3), (0,0.1,0.2,0.2)), ((0.4,0.5,0.6,0.7), (0.0,0.1,0.2,0.3), (0.1,0.1,0.1,0.1))] |
| FR | [((0.4,0.5,0.6,0.7), (0.0,0.1,0.2,0.3), (0.1,0.1,0.1,0.1)), ((0.7, 0.7,0.7,0.7), (0.0,0.1,0.2,0.3), (0.1,0.1,0.1,0.1))] | [((0.0,0.1,0.1,0.2), (0.1,0.1,0.1,0.1), (0.6,0.7,0.8,0.9)), ((0.7, 0.7,0.7,0.7), (0.0,0.1,0.2,0.3), (0.1,0.1,0.1,0.1))] | [((0.0,0.1,0.1,0.2), (0.1,0.1,0.1,0.1), (0.6,0.7,0.8,0.9)), ((0.4,0.5,0.6,0.7), (0.0,0.1,0.2,0.3), (0.1,0.1,0.1,0.1))] | [((0.2,0.3,0.4,0.5), (0,0.1,0.2,0.3), (0,0.1,0.2,0.2)), ((0.4,0.5,0.6,0.7), (0.0,0.1,0.2,0.3), (0.1,0.1,0.1,0.1))] | [((0.4,0.5,0.6,0.7), (0.0,0.1,0.2,0.3), (0.1,0.1,0.1,0.1)), ((0.7, 0.7,0.7,0.7), (0.0,0.1,0.2,0.3), (0.1,0.1,0.1,0.1))] |
| IR | [((0.2,0.3,0.4,0.5), (0,0.1,0.2,0.3), (0,0.1,0.2,0.2)), ((0.7, 0.7,0.7,0.7), (0.0,0.1,0.2,0.3), (0.1,0.1,0.1,0.1))] | [((0.4,0.5,0.6,0.7), (0.0,0.1,0.2,0.3), (0.1,0.1,0.1,0.1)), ((0.7, 0.7,0.7,0.7), (0.0,0.1,0.2,0.3), (0.1,0.1,0.1,0.1))] | [((0.2,0.3,0.4,0.5), (0,0.1,0.2,0.3), (0,0.1,0.2,0.2)), ((0.4,0.5,0.6,0.7), (0.0,0.1,0.2,0.3), (0.1,0.1,0.1,0.1))] | [((0.2,0.3,0.4,0.5), (0,0.1,0.2,0.3), (0,0.1,0.2,0.2)), ((0.7, 0.7,0.7,0.7), (0.0,0.1,0.2,0.3), (0.1,0.1,0.1,0.1))] | [((0.4,0.5,0.6,0.7), (0.0,0.1,0.2,0.3), (0.1,0.1,0.1,0.1)), ((0.7, 0.7,0.7,0.7), (0.0,0.1,0.2,0.3), (0.1,0.1,0.1,0.1))] |
| SM | [((0.2,0.3,0.4,0.5), (0,0.1,0.2,0.3), (0,0.1,0.2,0.2)), ((0.7, 0.7,0.7,0.7), (0.0,0.1,0.2,0.3), (0.1,0.1,0.1,0.1))] | [((0.2,0.3,0.4,0.5), (0,0.1,0.2,0.3), (0,0.1,0.2,0.2)), ((0.7, 0.7,0.7,0.7), (0.0,0.1,0.2,0.3), (0.1,0.1,0.1,0.1))] | [((0.4,0.5,0.6,0.7), (0.0,0.1,0.2,0.3), (0.1,0.1,0.1,0.1)), ((0.7, 0.7,0.7,0.7), (0.0,0.1,0.2,0.3), (0.1,0.1,0.1,0.1))] | [((0.4,0.5,0.6,0.7), (0.0,0.1,0.2,0.3), (0.1,0.1,0.1,0.1)), ((0.7, 0.7,0.7,0.7), (0.0,0.1,0.2,0.3), (0.1,0.1,0.1,0.1))] | [((0.0,0.1,0.1,0.2), (0.1,0.1,0.1,0.1), (0.6,0.7,0.8,0.9)), ((0.4,0.5,0.6,0.7), (0.0,0.1,0.2,0.3), (0.1,0.1,0.1,0.1))] |
| M M | [((0.0,0.1,0.1,0.2), (0.1,0.1,0.1,0.1), (0.6,0.7,0.8,0.9)), ((0.7, 0.7,0.7,0.7), (0.0,0.1,0.2,0.3), (0.1,0.1,0.1,0.1))] | [((0.0,0.1,0.1,0.2), (0.1,0.1,0.1,0.1), (0.6,0.7,0.8,0.9)), ((0.4,0.5,0.6,0.7), (0.0,0.1,0.2,0.3), (0.1,0.1,0.1,0.1))] | [((0.0,0.1,0.1,0.2), (0.1,0.1,0.1,0.1), (0.6,0.7,0.8,0.9)), ((0.7, 0.7,0.7,0.7), (0.0,0.1,0.2,0.3), (0.1,0.1,0.1,0.1))] | [((0.2,0.3,0.4,0.5), (0,0.1,0.2,0.3), (0,0.1,0.2,0.2)), ((0.4,0.5,0.6,0.7), (0.0,0.1,0.2,0.3), (0.1,0.1,0.1,0.1))] | [((0.4,0.5,0.6,0.7), (0.0,0.1,0.2,0.3), (0.1,0.1,0.1,0.1)), ((0.7, 0.7,0.7,0.7), (0.0,0.1,0.2,0.3), (0.1,0.1,0.1,0.1))] |
| CK | [((0.0,0.1,0.1,0.2), (0.1,0.1,0.1,0.1), (0.6,0.7,0.8,0.9)), ((0.7, 0.7,0.7,0.7), (0.0,0.1,0.2,0.3), (0.1,0.1,0.1,0.1))] | [((0.4,0.5,0.6,0.7), (0.0,0.1,0.2,0.3), (0.1,0.1,0.1,0.1)), ((0.7, 0.7,0.7,0.7), (0.0,0.1,0.2,0.3), (0.1,0.1,0.1,0.1))] | [((0.4,0.5,0.6,0.7), (0.0,0.1,0.2,0.3), (0.1,0.1,0.1,0.1)), ((0.7, 0.7,0.7,0.7), (0.0,0.1,0.2,0.3), (0.1,0.1,0.1,0.1))] | [((0.0,0.1,0.1,0.2), (0.1,0.1,0.1,0.1), (0.6,0.7,0.8,0.9)), ((0.4,0.5,0.6,0.7), (0.0,0.1,0.2,0.3), (0.1,0.1,0.1,0.1))] | [((0.2,0.3,0.4,0.5), (0,0.1,0.2,0.3), (0,0.1,0.2,0.2)), ((0.4,0.5,0.6,0.7), (0.0,0.1,0.2,0.3), (0.1,0.1,0.1,0.1))] |

Further we implement the proposed method to find the best alternative to achieve the authentication system:

*Step1:* Implement the IVTrNWAA operator to get the combined IVTrNNs in the form of $\tilde{\tilde{d}}_i (i = 1, 2, 3, \ldots, 5)$ for each alternative $A_i (i = 1, 2, 3, \ldots, 8)$ as shown in Table IV.

TABLE IV: COMBINED INTERVAL VALUED TRAPEZOIDAL NEUTROSOPHIC NUMBERS

| $A_i$ ($i = 1, 2, 3, \ldots, 8$) | $\tilde{\tilde{d}}_i (i = 1, 2, 3, \ldots, 5)$ |
|---|---|
| PW | [((0.2555,0.3732,0.4719,0.5838), (0,0.0562,0.1125,0.1687), (0,0.0915,0.1591,0.1638)), ((0.6967,0.7206,0.7473,0.7780), (0,0.0562,0.1337,0.2220), (0.0562, 0.0562, 0.0562, 0.0562))] |
| TF | [((0.2240,0.3437,0.4273,0.5437), (0,0.0562,0.0946,0.1282), (0,0.1488, 0.2173, 0.2305)), ((0.6880,0.7134,0.7436,0.7780), (0,0.0562,0.1337,0.2220), (0.0562, 0.0562, 0.0562, 0.0562))] |
| CT | [((0.1676,0.2893,0.3533,0.4736), (0,0.0562,0.0795,0.0974), (0, 0.2420,0.3181,0.3475)), ((0.7360,0.7477,0.7614,0.7780), (0,0.0562,0.1337,0.2220), (0.0562, 0.0562, 0.0562, 0.0562))] |
| FR | [((0.2592,0.3786,0.4611,0.5783), (0,0.0562, 0.0914, 0.1213), (0,0.1640,0.2027,0.2163)), ((0.6860,0.7134,0.7436,0.7780), (0,0.0562,0.1337,0.2220), (0.0562, 0.0562, 0.0562, 0.0562))] |
| IR | [((0.3448,0.4589,0.5688,0.6743), (0,0.0562,0.1337,0.2220), (0.0562, 0.0562, 0,0.0946, 0.0946)), ((0.7360,0.7477,0.7614,0.7780), (0,0.0562,0.1337,0.2220), (0.0562, 0.0562, 0.0562, 0.0562))] |
| SM | [((0.3072,0.4238,0.5228,0.6337), (0,0.0562,0.1125,0.1687), (0,0.0915,0.1337,0.1377)), ((0.7360,0.7477,0.7614,0.7780), (0,0.0562,0.1337,0.2220), (0.0562, 0.0562, 0.0562, 0.0562))] |
| MM | [((0.1583,0.2803,0.3400,0.4611), (0,0.0562,0.0768,0.0922), (0,0.2667,0.3409,0.3746)), ((0.6967,0.7206,0.7473,0.7780), (0,0.0562,0.1337,0.2220), (0.0562, 0.0562, 0.0562, 0.0562))] |
| CK | [((0.2859,0.4042,0.4929,0.6078), (0,0.0562,0.0979,0.1354), (0,0.1350,0.1705,0.1797)), ((0.6967,0.7206,0.7473,0.7780), (0,0.0562,0.1337,0.2220), (0.0562, 0.0562, 0.0562, 0.0562))] |

*Step2:* Calculate the score $S(\tilde{\tilde{d}}_i)(i = 1, 2, 3, \ldots, 8)$ to rank the alternatives $A_i (i = 1, 2, 3, \ldots, 8)$ as shown in table V:

TABLE V: SCORE OF EACH ALTERNATIVE

| $A_i$ ($i = 1, 2, 3, \ldots, 8$) | $S(\tilde{\tilde{d}}_i)$ ($i = 1, 2, 3, \ldots, 8$) |
|---|---|
| IR | 0.8232 |
| SM | 0.8156 |
| CK | 0.8051 |
| PW | 0.8016 |
| FR | 0.7962 |
| TF | 0.7895 |
| CT | 0.772 |
| MM | 0.7593 |

*Step3:* Based on the score values, alternatives are ranked as $IR \succ SM \succ CK \succ PW \succ FR \succ TF \succ CT \succ MM$. The symbol $\succ$ indicates "preferred to" and we can see that "Memory Card" is the desirable alternative to implement authentication as per the expert.

The proposed methodology can be applied to those MADM situations where (1) information is neutrosophic in the unit interval of real numbers and can be represented in the form of a trapezoid (2) information is neutrosophic in the unit interval of real numbers and can be represented in the form of a triangle. Whereas decision-making method proposed in [20] fits for the decision-making problems where (1) information is neutrosophic and can be represented in the form of a trapezoid (2) information is neutrosophic and can be represented in the form of a triangle.

Thus, the method proposed in the paper fits those situations where expert's opinion is in the range of acceptable behavior which was missing in [20].

## 9. CONCLUSION

The paper proposes an interval valued trapezoidal neutrosophic set(IVTrNS) which is generalization of IVIFN and TrNS. The paper also introduces the operational laws for the proposed IVTrNN. Further an IVTrNWAA operator is introduced to combine the trapezoidal information which is neutrosophic and in the unit interval of real numbers.

Finally, a method is developed to handle the problems in the multi attribute decision making(MADM) environment using IVTrNWAA operator followed by a numerical example of NFRs prioritization to illustrate the relevance of the developed method. The advantage of this method is that it helps in solving MADM problems where information is uncertain, inconsistent, imprecise or indeterminate; and it is between some range of acceptable behaviour. However, with respect to the proposed method, constructing IVTrNNs is a key problem to extract the truth, indeterminacy and falsity membership functions, whose values depend on both the intervals and trapezoidal fuzzy numbers.